\documentclass[conference]{IEEEtran}
\IEEEoverridecommandlockouts

\usepackage{cite}
\usepackage{amsmath,amssymb,amsfonts}
\usepackage{algorithmic}
\usepackage{graphicx}
\usepackage{flushend}
\usepackage[numbers,sort&compress]{natbib}
\usepackage[capitalise]{cleveref}
\usepackage{textcomp}
\usepackage{xcolor,colortbl}
\usepackage[top=54pt, left=54pt, right=54pt, bottom=54pt]{geometry}%
\def\BibTeX{{\rm B\kern-.05em{\sc i\kern-.025em b}\kern-.08em
    T\kern-.1667em\lower.2ex\hbox{E}\kern-.125emX}}

\definecolor{Gray}{gray}{0.75}
\definecolor{Grey}{gray}{0.92}

\newcolumntype{a}{>{\columncolor{Grey}}c}

\title{TumblerBots: Tumbling Robotic sensors for Minimally-invasive Benthic Monitoring}

\begin{document}

\author{L. Romanello$^{1,2}$, A. Teboul$^{1}$, F. Wiesemüller$^{2,3,4}$, P. H. Nguyen$^{3}$, M. Kovac$^{3,4}$, S. F. Armanini$^{2}$  \\

\thanks{ 
\par $^{1}$eAviation Laboratory, TUM School of Engineering and Design, Munich.
\par $^{2}$Aerial Robotics Laboratory, Imperial College London.
\par $^{3}$Laboratory of Sustainability Robotics, EMPA, Dübendorf, Switzerland.
\par $^{4}$ Laboratory of Sustainability Robotics, EPFL, Lausanne Switzerland.}
}

\maketitle
\begin{abstract}
    Robotic systems show significant promise for water environmental sensing applications such as water quality monitoring, pollution mapping and biodiversity data collection. 
    Conventional deployment methods often disrupt fragile ecosystems, preventing depiction of the undisturbed environmental condition. In response to this challenge, we propose a novel framework utilizing a lightweight tumbler system equipped with a sensing unit, deployed via a drone. This design minimizes disruption to the water habitat by maintaining a slow descent. The sensing unit is detached once on the water surface, enabling precise and non-invasive data collection from the benthic zone. 
    The tumbler is designed to be lightweight and compact, enabling deployment via a drone. The sensing pod, which detaches from the tumbler and descends to the bottom of the water body, is equipped with temperature and pressure sensors, as well as a buoyancy system. The later, activated upon task completion, utilizes a silicon membrane inflated via a chemical reaction. The reaction generates a pressure of 70 kPa, causing the silicon membrane to expand by 30\%, which exceeds the 5.7\% volume increase required for positive buoyancy. The tumblers, made from ecofriendly materials to minimize environmental impact when lost during the mission, were tested for their gliding ratio and descent rate. They exhibit a low descent rate, in the range of 0.8 to 2.5 meters per seconds, which minimizes disturbance to the ecosystem upon water landing. Additionally, the system demonstrated robustness in moderate to strong wind conditions during outdoor tests, validating the overall framework.
 
\end{abstract}


\section{Introduction}

Aquatic environments are invaluable ecosystems, supporting biodiversity ecosystems' essential role in preserving ecological balance and contributing to economic prosperity worldwide \cite{strayer2010freshwater}. Aquatic environments, which cover 71\% of the Earth's surface, sustain at least 25\% of all known species, while freshwater environments account for approximately 10\% of global biodiversity. However, much of this critical resource remains inaccessible, as significant portions are located in volcanic lakes, glaciers and deep forest water sources, making comprehensive study and effective management challenging. Human activity and the escalating threat of climate change further endanger these ecosystems. Alarmingly, recent studies have shown that freshwater species populations have declined by 84\%, emphasizing the urgent need for conservation efforts \cite{wwf2020living}. From 1970 to 2015, natural wetlands globally declined at an average annual rate of -0.95\% per year, with the rate nearly doubling to -1.6\% per year from 2010 to 2015, a pace three times faster than the rate of forest loss (-0.24\% per year, 1990-2010). \cite{darrah2019wetland}.
\begin{figure}[t]
\centering
\includegraphics[width=1\columnwidth]{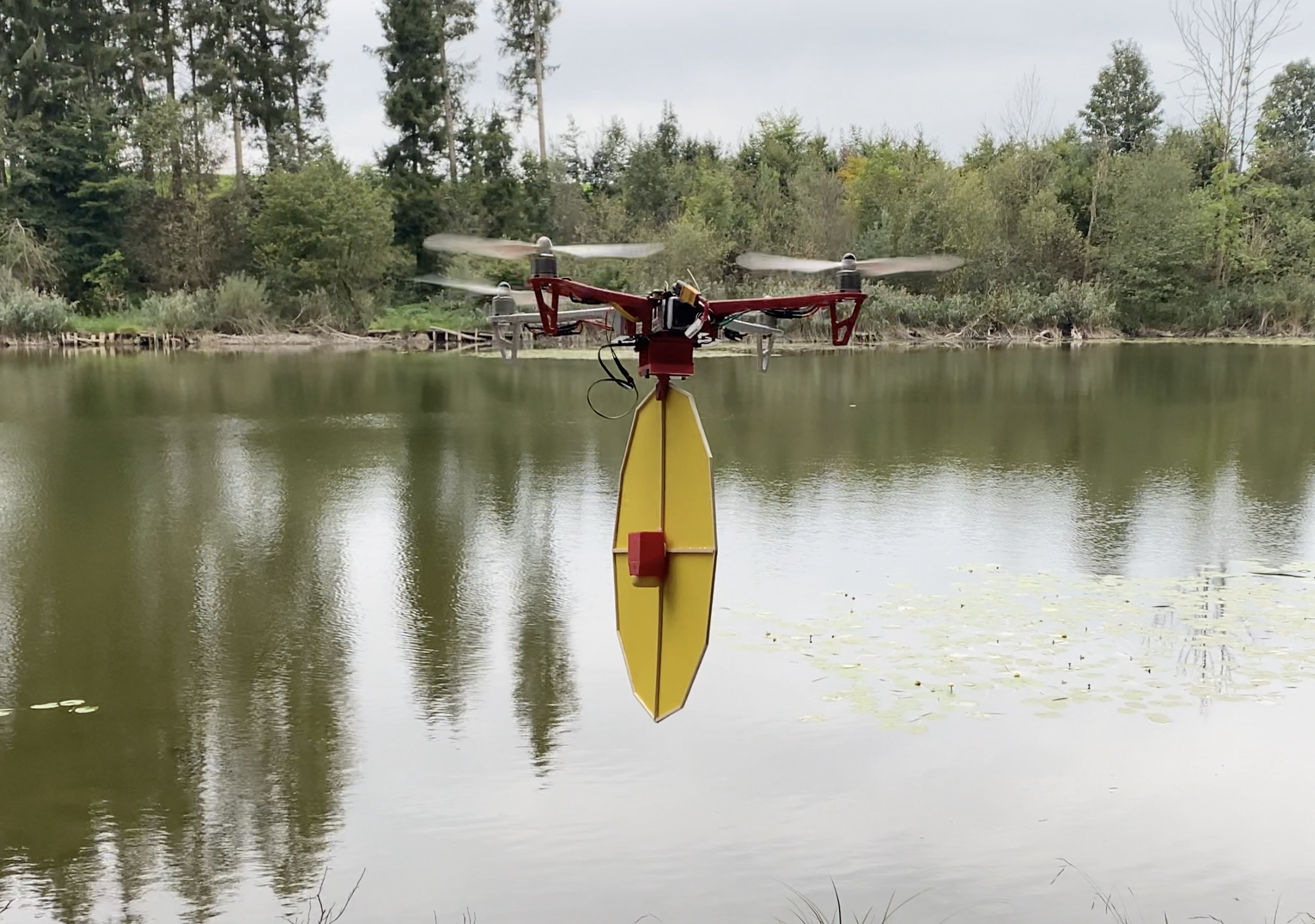}
\caption{The drone, holding and ready to deploy the tumbler (TumblerPod), during an outdoor mission over a lake.}
\label{fig:glider_1st}
\end{figure}
A major challenge in addressing these threats is the lack of comprehensive data, particularly in remote or hazardous areas. Biologists' aquatic missions typically rely on large vessels and divers, which involve risks, automation challenges, and high costs.
\begin{figure*}[t!]
\centering
\includegraphics[width=0.98\textwidth]{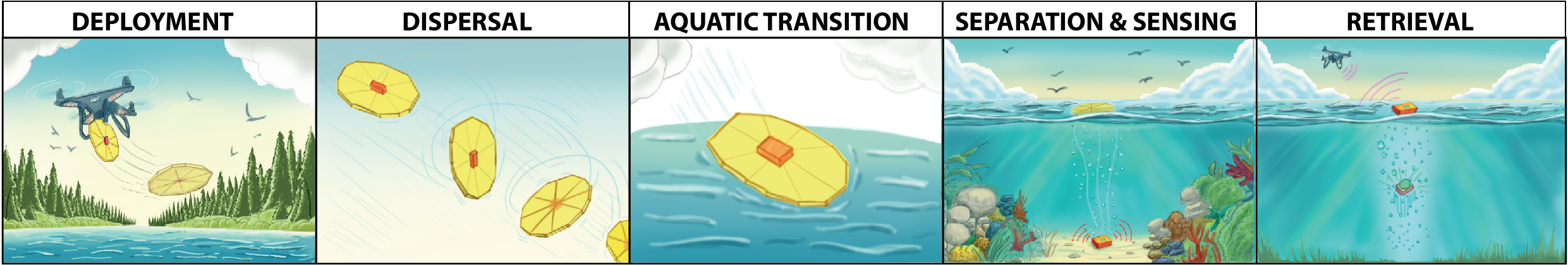}
\caption{Overall mission [From Left to Right]: The drone flies to deployment location above the water and releases the tumbler (TumblerPod). The tumbler steadily tumbles towards the water surface. The sensor pod then breaks away and sinks to the bottom of the lake and starts taking ecological measurements of the region. After it finishes its task, the buoyancy module activates through chemical reaction and the pod rises to the surface of the lake as a retrieval drone arrives to pick it up.}
\label{fig:mission}
\vspace{-1.5em}
\end{figure*}
Robotic systems present a solution for monitoring aquatic environments, mitigating the hazards faced by divers during specimen retrieval. These systems are integral to a wide range of underwater operations, such as the recovery of floating, submerged, or seabed items, as well as biological sample collection \cite{galloway2016soft}. By enhancing operational safety, increasing efficiency, and reducing overall mission costs, robotic tools have become indispensable in underwater exploration \cite{gong2021soft, doi:10.1126/scirobotics.aaz1012, giordano2024soft}. 
Typically employed as Remotely Operated Vehicles (ROVs), these systems are often large-scale, posing logistical challenges in transportation and deployment from surface vessels \cite{casalino, doi:10.34133/cbsystems.0089}. Moreover, their size and operational complexity can lead to significant disturbances to marine life and introduce environmental risks. In fact, sunken boats and ROVs become environmental waste on the ocean floor in the event of connection failures or vehicle damage. Furthermore, the reliance on robots using a single mode of locomotion limits their access to remote or hazardous locations, including mountainous regions, volcanic lakes, and forest water sources.
A viable approach to addressing this challenge is the development of systems combining flight and underwater operation capabilities \cite{Siddall2014,rockenbauer2021dipper,zufferey2019consecutive,maia2017demonstrating}. A notable contribution in this domain is the work by Debruyn et al. \cite{debruyn2020medusa}, which presents a multi-rotor platform capable of landing on water and deploying an underwater pod for sensing missions. However, these systems are incapable of performing sampling tasks and benthic sensing —measurements conducted on the ocean floor— and their propulsion mechanisms, many involving abrupt water-air transitions \cite{zufferey2019consecutive, Siddall2014}, often result in environmental disturbances.
This work aims to minimize disturbances during water entry as well as collisions underwater, while collecting key water quality data, such as salinity, pH, temperature, and turbidity. These capabilities are crucial for remote areas and benthic zone exploration, where sensitive data collection with minimal ecological impact is essential.
A potential solution arises from the concept of biogliders \cite{Sethi2022, kim2021microfliers, 7054018}, which offer a sustainable, non-invasive approach to forest sensing. Examples include biodegradable gliders equipped with pH sensors that detect environmental changes through color alteration \cite{10.3389/frobt.2022.1011793}. Gliders have demonstrated long-term effectiveness in remote, non-invasive ecosystem monitoring and research \cite{s21206752, testor2019oceangliders, kovac2009selfdeploying}, with the advantage of covering large areas with minimal energy expenditure. However, deploying gliders for aquatic data-collection introduces additional challenges, such as a high impact velocity when entering water. 
An alternative solution that avoids the problem of high water impact velocity, are parachutes. Parachutes deployed for sensor placement offer a high payload capacity and compact storage, enabling sensors to be placed in hard-to-reach areas for environmental monitoring \cite{nasa_parachute}. However, successful and accurate deployment is strongly affected by wind and weather conditions. 
Inspired by the descending maneuvers of leaves and underwater robotic sheets \cite{JunghwanByun.2021}, we propose a tumbling system design integrated with a water sensing unit. For this sensing pod to function effectively in an ecological context, integrating electronics is essential, even though these components are not biodegradable.
The system is deployed by a drone and released over water bodies such as lakes, where it descends smoothly towards the water surface. Upon water entry, the sensing unit reaches the benthic zone with minimal hydrodynamic disturbance, enabling effective data collection in the aquatic environment. \\
Our contributions include: 
\begin{itemize} 
    \item Design and development methodology for a novel framework for non-invasive benthic monitoring.
    \item First-time demonstration of robotic tumbling systems equipped with varying payloads.
    \item Investigation of geometrical parameters through exploration of different tumbling geometries and structures. 
    \item Exploration and validation of an innovative buoyancy system using a soft bladder inflated via chemical reaction.
    \item Validation of the proposed framework for safe operation in outdoor aquatic environments.
\end{itemize}

The development of safer and more eco-friendly sensing systems has the potential to enable more sustainable approaches to environmental data acquisition.



\section{Methodology \& Materials}

The framework is composed of three main components, each with specific system requirements according to the mission objectives:
\begin{itemize}
    \item \textit{Drone and Gripper}: A commercially available small-scale UAV should be selected to minimize costs and improve accessibility for biologists. The gripper must be simple and lightweight, minimizing the payload to optimize the drone's flight performance.
    \item \textit{Tumbler}: The tumbler must be biodegradable and achieve a low descent rate, with limited drift from the deployment point. It should also be simple to design, easy to manufacture, and scalable to accommodate different sensing units. It must be resistant to external factors such as weather conditions and wind during descent.
    \item \textit{Sensing Pod}: The sensing pod requires the integration of essential electronics and a power source for ecological data collection. Although these components are non-biodegradable, they are necessary for the system's functionality in environmental monitoring. A buoyancy mechanism must also be integrated in order for the sensing pod to rise to the surface at the end of the mission.
\end{itemize}

\subsection{Mission}
The drone navigates toward the designated water body. Upon reaching a safe altitude above the target area — ensuring minimal disruption to local aquatic species — the tumbler(s) is released. As the tumbler descends, tumbling at a moderate speed - between 1 and 3 m/s -, it approaches the water surface. Upon contact, a detachable module (i.e. the Benthic Sensor Pod), equipped with sensing capabilities, data storage, a localization system and a buoyancy system, separates and sinks to the bottom to conduct monitoring and sensing operations. The remaining glider component remains on the surface and is left to biodegrade over time in the environment. The mission duration varies depending on the specific task. Upon completion, or when triggered, the buoyancy system of the pod is activated, causing the device to ascend to the surface. The pod's localization system allows for retrieval by either a drone or a watercraft, or for collection upon drifting to shore. Data gathered during the mission is stored in a memory card for later analysis. 

\subsection{Tumbler}
\subsubsection{Tumbler choice}

The choice of such an unconventional deployment system stems from the unique mission requirements. The deployment system should not only ensure a slow descent rate but also be lightweight, biodegradable, and easy to manufacture. While biogliders have previously found use in environmental monitoring \cite{10.3389/frobt.2022.1011793}, in order to have more onboard sensors, the wingspan of the bioglider would need to be increased. Indeed, designing a suitable glider for our payload mass - 70 g - would require increasing the wingspan significantly to a length greater than what the carrier drone allows. Additionally, the 
added payload increases the manufacturing complexity of the glider due to the need for a range of stiffeners such as stringers, ribs, and spars to maintain adequate wing strength. Moreover, as they cannot compensate for imbalances, traditional winged gliders are sensitive to payload changes; they rely on a precise balance between their center of gravity (CG) and the center of pressure (CP) for stable flight \cite{Li.2022}. This, in turn, means that any change in the sensor node will have a detrimental impact on performance unless design changes are carried out. 

Another possible deployment method comes in the form of parachutes, which at first glance offer numerous advantages. They not only have a high payload carrying capacity but can be folded into a small, lightweight package. However, parachutes are very sensitive to wind patterns \cite{Pham.2022}, which limits the practicability of the system, as calm conditions would be required to ensure the sensing units land on the water.
Tumbling systems are not as affected by wind patterns, yet they can provide seamless water landing task due to their low descend rates. Crucially, due to the simple geometry of tumbling structures, they are lightweight and easy to manufacture even with biodegradable materials such as cellulose.

\subsubsection{Tumbler dynamics}
Tumbling is one of multiple falling regimes of flat objects, including steady-falling, fluttering, and chaotic motion \cite{Field.1997, Auguste.2013}. First explored by Willmarth in 1964, the passive fall dynamics of discs is governed by intrinsic parameters of the moving object and surrounding fluid, such as body geometry, body-to-fluid density ratio (r/rf), and Reynolds number (Re) \cite{Willmarth.1964}. Most importantly, the dimensionless moment of inertia of the disk was identified as a key parameter:\\
\[I^* = \frac{I}{\rho_1 d^5} = \frac{\pi \rho_2 t}{64 \rho_1 d}\]\\
It is formed from the ratio of the moment of inertia of a thin disk about a diameter and a quantity proportional to the moment of inertia of a rigid sphere of fluid about its diameter, \(d\). This parameter arises naturally in the inertial terms of the equations of motion for the rotation of the disk about the mass center, thereby influencing the type of falling regime. Indeed, for low \(I^*\) values, where the moment of inertia of the disc is negligible, objects tend to flutter, akin to the side by side oscillation of falling leaves. While fluttering could further reduce descend rates, designing a thin disc with a sufficiently low \(I^*\) to flutter in air is extremely challenging, even before accounting for the payload.

Tumbling sets in for higher \(I^*\) values and remains the dominant fall regime for a wide range of \(I^*\) values,from 0.04 to at least 0.2 \cite{Field.1997}. One advantage of tumbling is its stability, meaning that the deployment system should work for a variety of payloads, as discussed later in Section \ref{sec:Tumbling_tests}.
Based on the main driving factors behind tumbling motion, the aim when designing the tumbler is to create a thin, lightweight, and stiff disc.

\subsubsection{Tumbler design} 
In this work, we utilize empirical prototyping to test our different tumbler designs, as the falling behavior of tumbling systems is extremely challenging to model analytically \cite{Howison.2020}.
Hence, different tumbler designs were constructed and iteratively refined. Design parameters include tumbler shape, cellulose sheet density, number of layers, and reinforcement structure pattern. The diameter of the disc itself is limited by the carrier drone which allows for a diameter of \( 40 cm\).

\begin{figure}[h]
    \centering
    \begin{minipage}{0.32\linewidth}
        \centering
        \includegraphics[width=\textwidth]{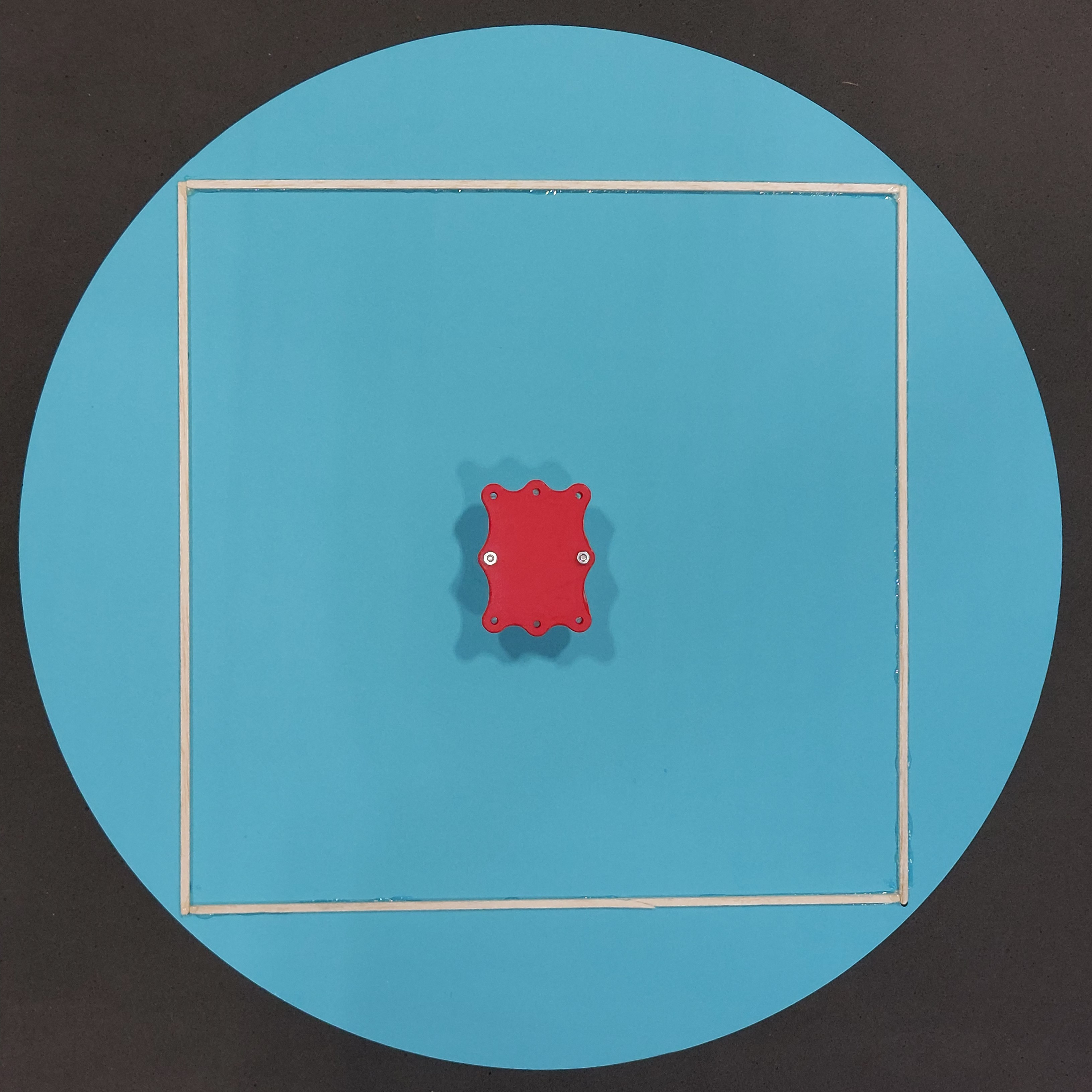}
    \end{minipage}%
    \hspace{0.003\linewidth} 
    \begin{minipage}{0.32\linewidth}
        \centering
        \includegraphics[width=\textwidth]{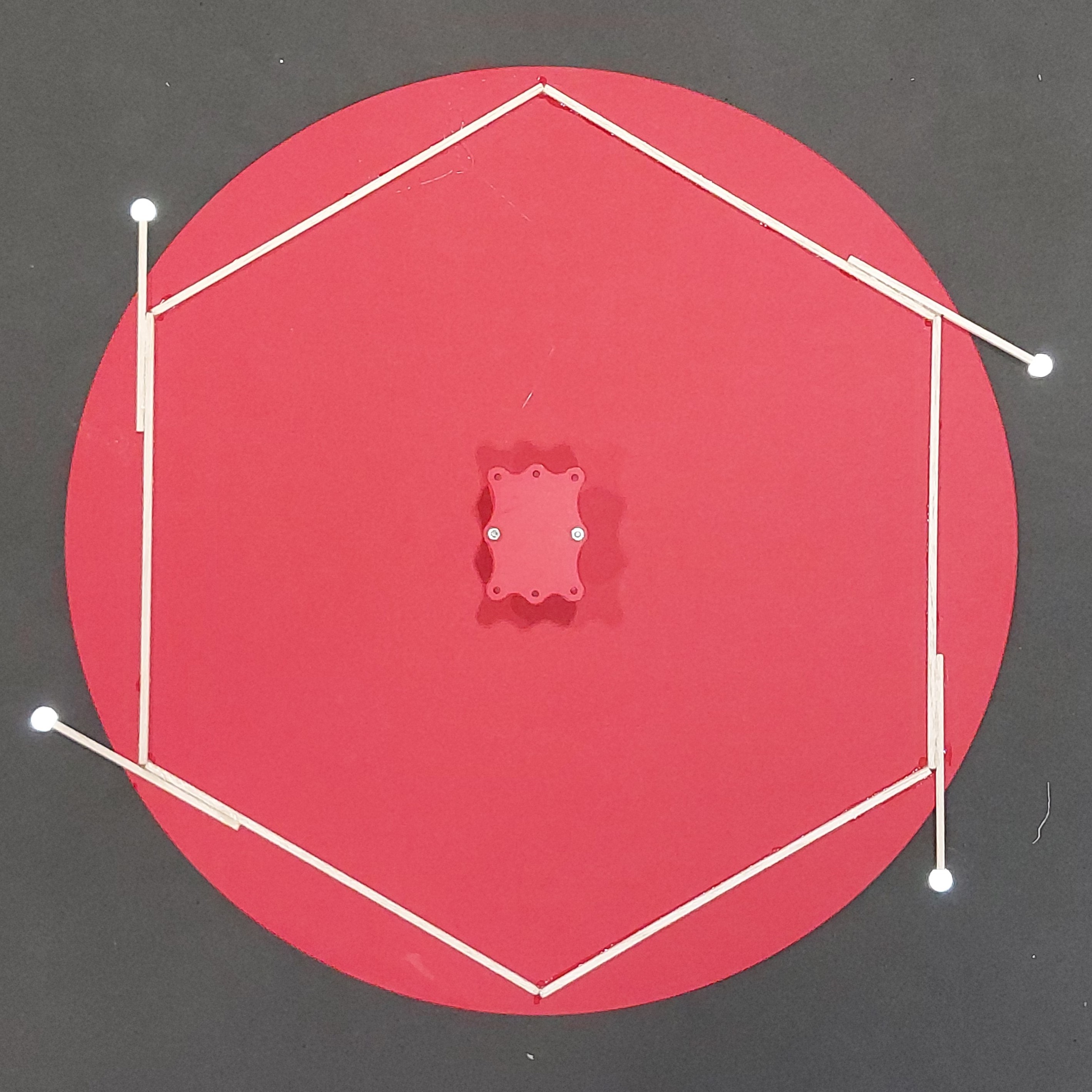}
    \end{minipage}
    \hspace{0.003\linewidth}
    \begin{minipage}{0.32\linewidth}
        \centering
        \includegraphics[width=\textwidth]{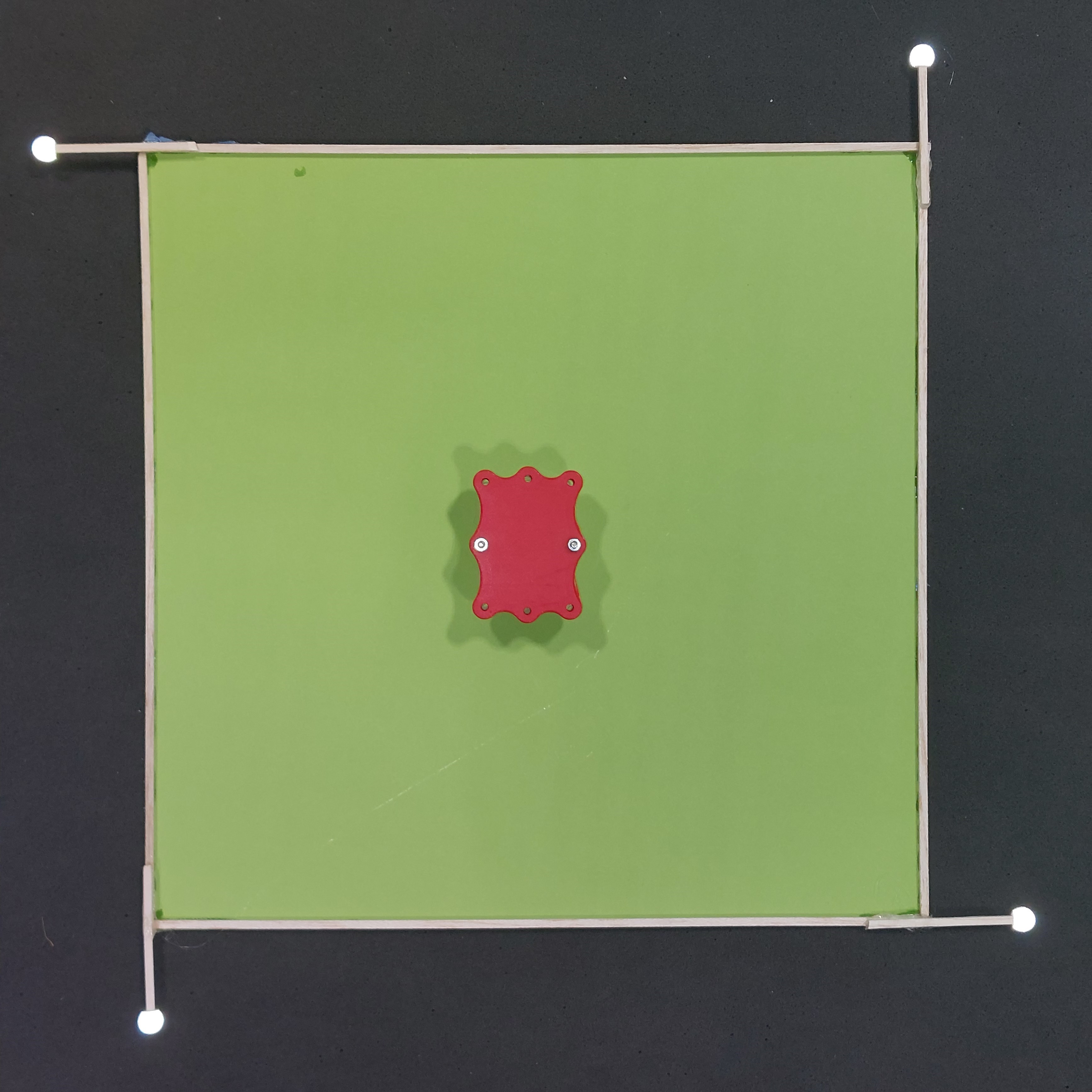}
    \end{minipage}%
    
    \vspace{0.5em} 
    
    \begin{minipage}{0.32\linewidth}
        \centering
        \includegraphics[width=\textwidth]{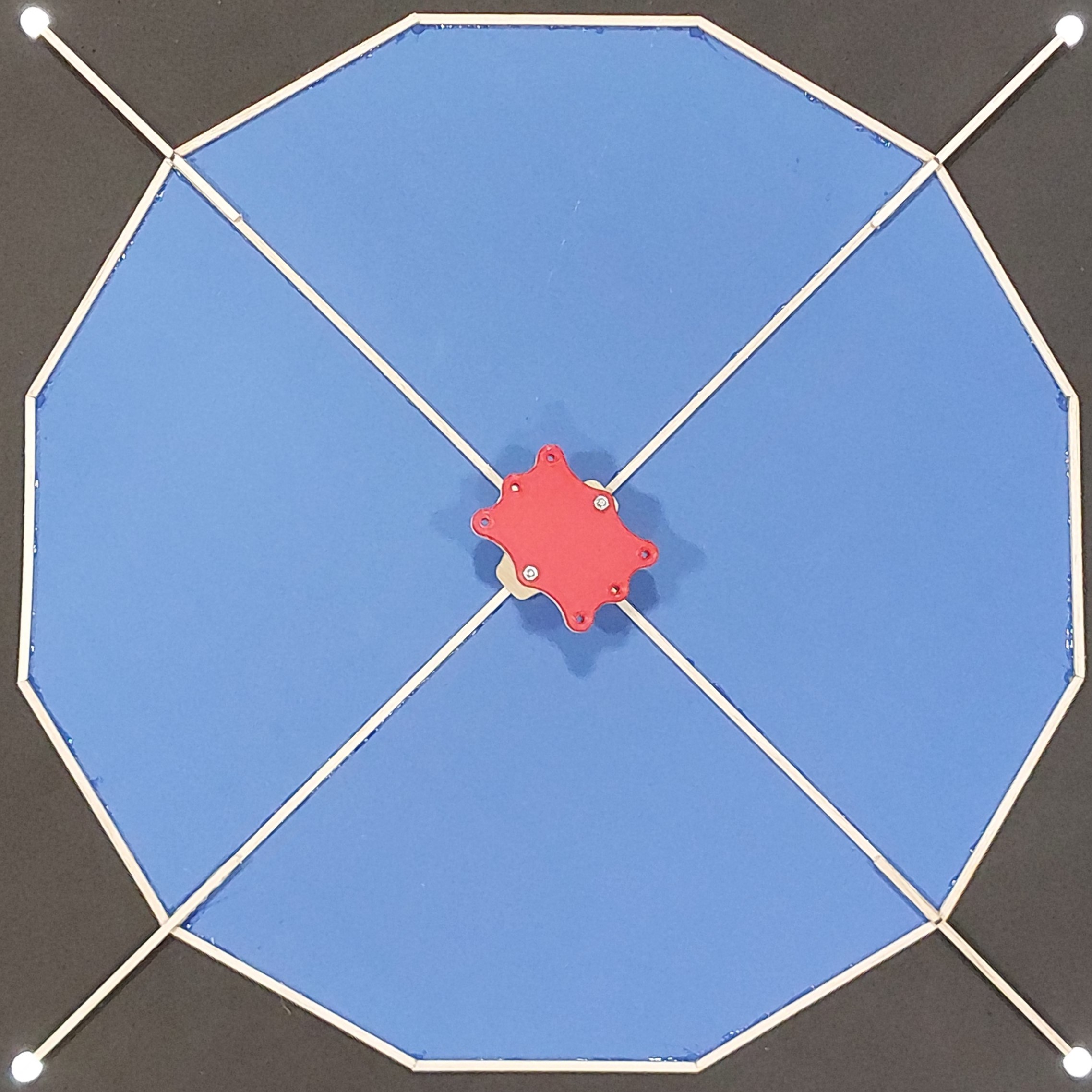}
    \end{minipage}
    \hspace{0.003\linewidth}
    \begin{minipage}{0.32\linewidth}
        \centering
        \includegraphics[width=\textwidth]{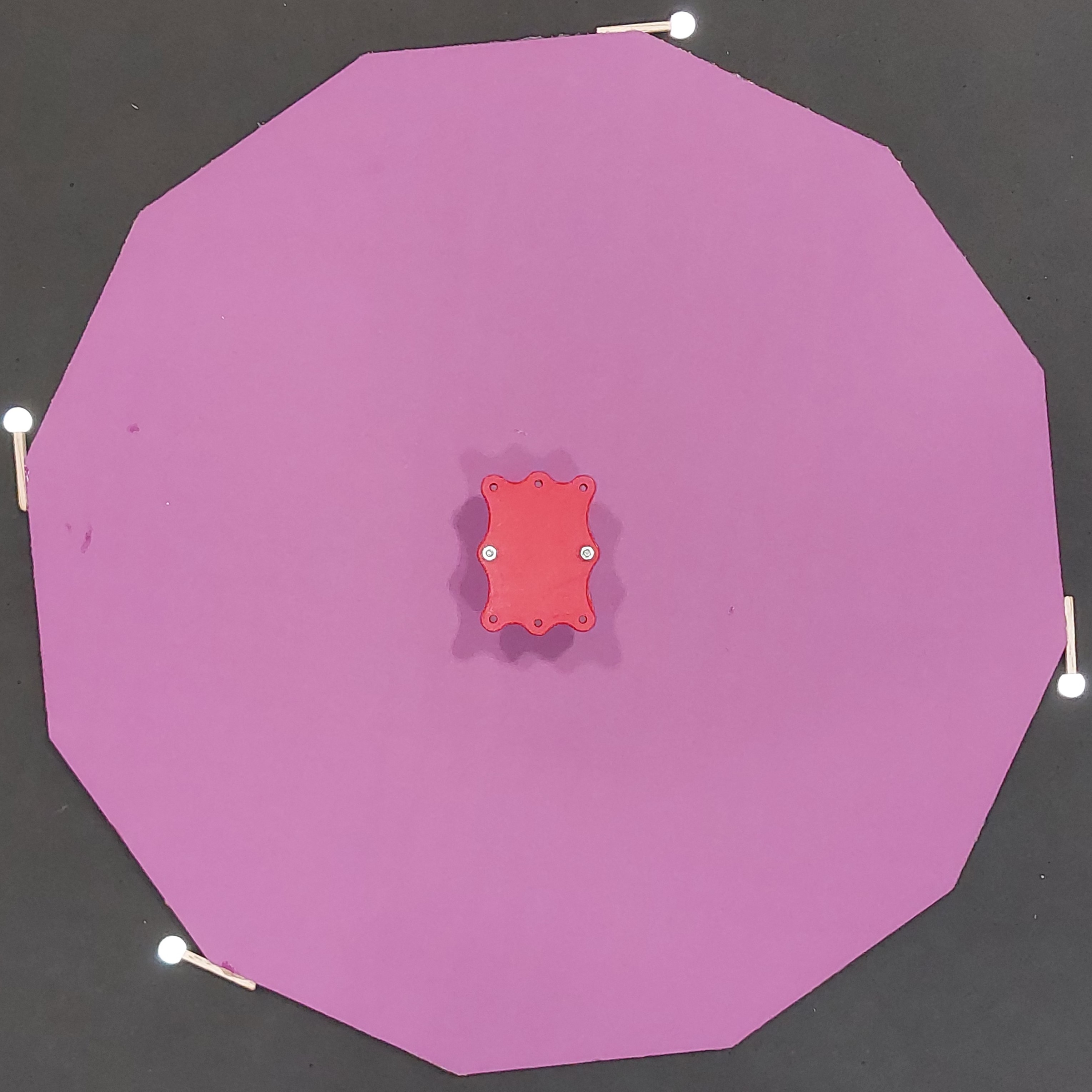}
    \end{minipage}%
    \hspace{0.003\linewidth} 
    \begin{minipage}{0.32\linewidth}
        \centering
        \includegraphics[width=\textwidth]{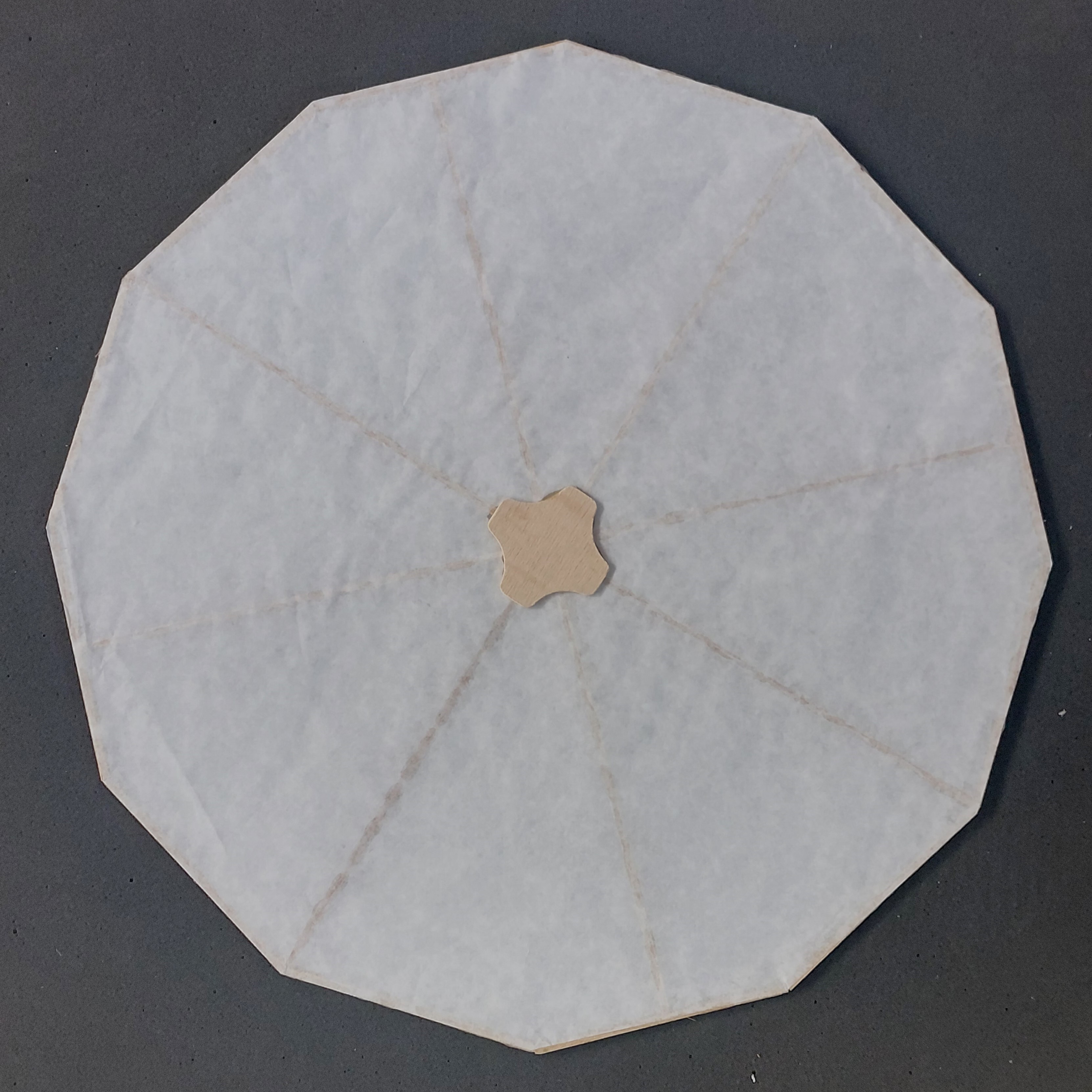}
    \end{minipage}

    \caption{Tumbler designs evaluated. From left to right, top to bottom: Circle1 (R, SL, HP, 32g), Circle2 (R, SL, HP, 33g), Square (R, SL, HP, 28g), Dodecagon1 (R, SL, HP, 36g), Dodecagon2 (R, DL, HP, 68g), Dodecagon3 (R, DL, LP, 22g).
R = reinforced / SL = single layer / DL = double layer/ HP = heavy paper / LP = light paper }
    \label{fig:tumbler_designs}
\end{figure}

From tests with simple discs of varying sizes cut from a thick cellulose sheet (Intenga, Saarland, Germany) with a density of 240 \(g/m^2\), it became apparent that for larger diameters of \(>\) 15cm, bending of the discs under the aerodynamic loads led to a chaotic fall regime. Accordingly, stiffeners in the form of 3mm square balsa sticks were added, which allowed for a more reliable tumbling motion. As seen in Circle1 and Circle2 in \cref{fig:tumbler_designs}, multiple reinforcement patterns were tested, with reinforcements closer to the leading edge proving key in improving tumbling behavior. 
In order to fully strengthen the leading edge with a straight stiffener, thus reducing any bending effects that might occur at the leading edge flap, we considered square-shaped tumblers, as seen in Square design of \cref{fig:tumbler_designs}. However, square tumblers proved less effective as they presented a noticeable lateral drift. It was hypothesized that the low order of rotational symmetry led to a periodic switch of the tumbling axis, leading to sharp changes in the dimensionless moment of inertia and therefore the fall dynamics. To avoid this problem, we opted for shapes with a higher order of rotational symmetry such as a dodecagon, as seen in further designs of \cref{fig:combined_figures}. While a simple dodecagon sheet with spar and leading edge reinforcement proved effective (Dodecagon1 design), its payload-carrying capacity was limited. A sandwich structure (Dodecangon2 design) was developed to improve stiffness, with two layers of cellulose sheets around the stiffeners. This design demonstrated highly reliable tumbling behavior, albeit with a low gliding ratio. The total mass is 68g, with 60g contributed by the dense cellulose sheets. Because the thickness of the cellulose sheet is not a significant contributing factor to the overall stiffness of the sandwich structure, the final design (Dodecagon3 design) uses a much lighter cellulose sheet (Eles Vida, Bremen, Germany) with a density of merely 40 \(g/m^2\), reducing the mass by a factor of 3 while retaining adequate stiffness, as characterized during experiments \cref{sec:Tumbling_tests}. An overview of the different tumbler designs is presented in \cref{fig:combined_figures}.

\subsection{Benthic Sensing pod}
\begin{figure}
\centering
\includegraphics[width=1\columnwidth]{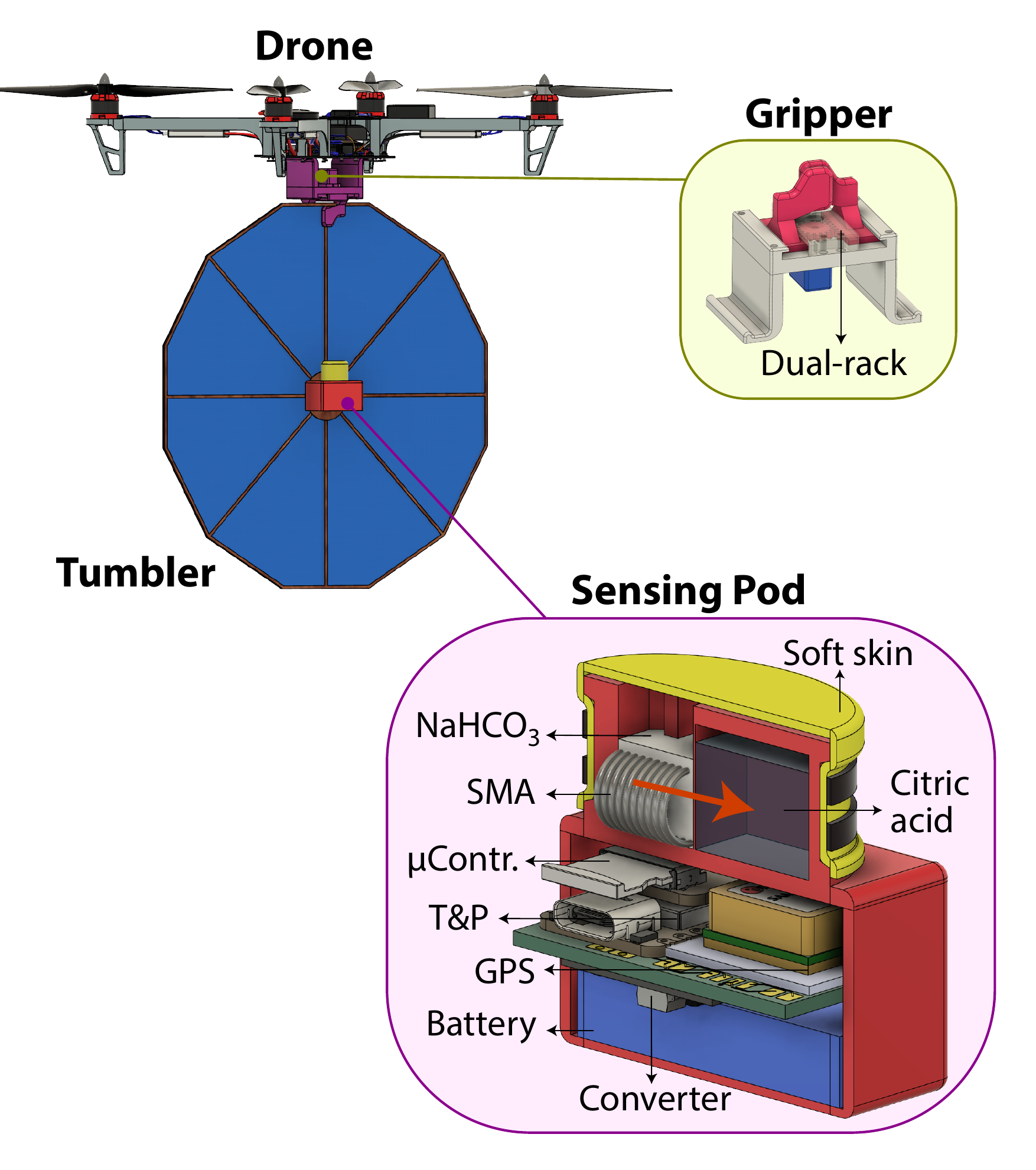}
\caption{(a) Overall framework consisting of the drone, gripper, tumbler, and sensing unit, with a detailed view of the gripper mechanism. (b) Sensing unit with a detailed view of its internal components.}
\label{fig:overall_framework}
\end{figure}

The sensing unit is highlighted in \cref{fig:overall_framework}, and is equipped with both temperature and pressure monitoring capabilities, a GPS module for surface object retrieval and tracking, as well as a buoyancy mechanism. Its onboard microcontroller is capable of storing the temperature and pressure data. An onboard energy source, paired with a power converter, ensures the necessary operational capacity.

\subsubsection{Electronics}
To achieve the tumbling state, the benthic sensing pod requires a lightweight design. Indeed, as shown in \cref{sec:Tumbling_tests}, above a threshold of 120g the reliability of the tumbling motion reduces dramatically and fails to meet the mission requirements. To meet the size and weight constraints for aerial deployment, a custom PCB was implemented, along with miniaturized components. The Seeed Xiao ESP32S3 integrates data storage, while the GPS connects via serial communication, and the temperature and pressure sensors interface through I2C. A 1S battery powers all components via a 5V converter, and the buoyancy system is controlled by a MOSFET managed by the microcontroller. \cref{tab:components} details the electronic components, resulting in a total payload of 70g.

\begin{table}

\caption{Specifications of sensing unit's components.}
\begin{center}
\scalebox{0.97}{
\begin{tabular}{|l|l|}
\hline
\rowcolor{Gray}
{\textbf{Components}} & {{\textbf{Implementation}}} \\
\hline
{Frame} & {Custom PCB} \\
\hline
{Micro-controller} & {Seeed Xiao ESP32S3} \\
\hline
{GPS module} & {M10Q-5883}\\
\hline
{T\&P Sensor} & {MS5837-30A} \\
\hline
{DC Converter} &  {BEC 5 V}  \\
\hline
{Mosfet} &  {IRFZ34NPbF}\\
\hline
{Battery} &  {1S 800 mAH Lipo} \\
\hline
{Total mass} &  {70 g} \\
\hline
\end{tabular}}
\label{tab:components}
\end{center}
\end{table}

\subsubsection{Monitoring sensors}
In this work, the sensing pod is equipped with a temperature and pressure sensor (MS5837-30A, BlueRobotics, California) to demonstrate the system's functionality. This allows for mapping water temperature based on depth. The system's modular design supports the integration of various other sensors, such as those for oxygen, salinity, or turbidity. Additionally, cameras \cite{ho2022vision} and microphones \cite{lawson2023use} can be incorporated without significantly increasing system complexity.

\subsubsection{Buoyancy}
We developed an onboard buoyancy mechanism to enable the collection of the sensor after mission completion. Thanks to the sensor's ability to return to the water surface on its own, the probability of losing the sensing pod is drastically reduced. 
We employ a silicon bladder (A35 Silicone, Trollfactory, Germany) - a soft membrane - to vary the buoyancy underwater. This has been demonstrated in other applications \cite{debruyn2020medusa,10522045}, where the bladder was inflatable via air pressure or reshaped by a servo mechanism.
In our application, using servos and pumps is not only inefficient since we only require a one-time solution but also impracticable due to the strict mass constraints. Consequently, we utilize a simpler, lighter mechanism that inflates the soft silicone skin using gas produced from a chemical reaction. This approach significantly reduces energy consumption while minimizing disturbance to underwater species. Specifically, the reaction of citric acid with sodium bicarbonate is employed. This acid-base reaction yields sodium citrate, water, and carbon dioxide gas, represented by the following equation:

\[
\text{C}_6\text{H}_8\text{O}_7 (aq) + 3 \, \text{NaHCO}_3 (s) \rightarrow 
\]
\[
\text{Na}_3\text{C}_6\text{H}_5\text{O}_7 (aq) + 3 \, \text{CO}_2 (g) + 3 \, \text{H}_2\text{O} (l)
\]

The carbon dioxide produced during this reaction serves as the gas utilized to inflate the silicon material. 
To ensure that the reaction occurs only after the completion of the sensing task or is triggered after a set time or depth limit, it is essential to keep the two reactants separated. This is achieved through the design of two distinct chambers, each housing one of the substances, with a thin wall separating them, as depicted in \cref{fig:overall_framework}. A shape memory alloy (SMA) spring is employed to breach this wall, facilitating the reaction as shown in \cref{fig:overall_framework}. The SMA spring acts as an actuator; when a voltage is applied, it elongates, generating a force. Following its activation, the spring can return to its original position, allowing for potential reuse.

The SMA utilized in our system has been tested and can produce forces of up to 3 N. To control the activation of the SMA, a MOSFET has been integrated into the electronics, triggered by a digital signal from the microcontroller.

\subsection{Drone \& Gripper}

To deploy the tumbler and sensing pod, a commercial drone (F450 DJI, Beijing) was utilized due to its low cost and compact size (see \cref{fig:overall_framework}). The drone is equipped with DJI 2212 920KV motors, 9450 propellers, and a Speedybee F405 V3 flight controller. The drone has a mass of 850g and a payload of 450g.
To facilitate the deployment of the tumbler, a custom-built gripper was developed as shown in \cref{fig:overall_framework}. The gripper is designed to securely hold the tumbler throughout the mission. It employs a dual rack drive mechanism, wherein a servomotor rotates a gear that moves two parallel racks in opposite directions. Each rack is equipped with a one-part holder, as illustrated in \cref{fig:overall_framework}. 

It is important to note that, as expected, the presence of the attached tumbler affects the drone's dynamics. Since the tumbler is mounted laterally relative to the drone, the quadrotor's roll motion is significantly influenced, resulting in reduced roll control. However, this does not substantially impact the feasibility of the system's deployment.

\subsection{Materials \& Manufacturing}

We utilized cellulose sheets to create the biodegradable base structure, and wood sticks to add stiffness to the structure, as seen \cref{fig:overall_framework}. The tumbler is attached to the sensing unit using Polyvinyl alcohol (PVA). This substance dissolves with water, enabling the sensing unit to detach and reach the benthic zone of the water body. PVA is partly biodegradable under certain circumstances, such as in the presence of specific microbial activity, moisture, and oxygen, making it an environmentally friendly alternative to conventional glues \cite{C6RA05742J}.

To facilitate the manufacturing of the sensing unit structure and minimize weight, both the enclosure for the electronics and the buoyancy system are printed in one piece. This is made possible by integrating the components into the structure during the printing process. The 0.35mm sacrificial wall was developed to be both watertight and to break upon application of a small force, ensuring that the SMA spring would break it.
The sensing unit structure is fully 3D printed in PHA (ALPHA, ColorFabb, VT, USA). Polyhydroxyalkanoates (PHAs) are biodegradable bioplastics that degrade in various environments through microbial activity, making them suitable for ecological applications. Their degradation rate depends on polymer composition and environmental factors \cite{Dai2008}, though according to specification in freshwater, a degradation of 90\% within 56 days is guaranteed \cite{Yu.2024}. For longer missions, a PLA version of the structure can be used, which can survive underwater for prolonged periods. Finally, the inflatable skin is molded using silicone with a hardness of A30.
To develop an effective data collection system for gathering insightful information from aquatic environments, the integration of electronics remains essential, even though these components are not biodegradable. Current energy sources, such as LiPo batteries, are also crucial for system functionality and efficiency. The total cost for the proposed tumbling system and sensing pod is under \$75, facilitated by a rapid manufacturing process. This simplicity in production would allow for easy distribution and deployment of the proposed framework.


\section{Tests \& Results} 
\subsection{Tumbling tests}
\label{sec:Tumbling_tests}
\begin{figure}[b!]
    \centering
    \includegraphics[width=0.091\textwidth]{Fig/20240826_161735.jpg}
    \includegraphics[width=0.091\textwidth]{Fig/20240826_154454.jpg}
    \includegraphics[width=0.091\textwidth]{Fig/20240826_154709.jpg}
    \includegraphics[width=0.091\textwidth]{Fig/20240826_154732.jpg}
    \includegraphics[width=0.0895\textwidth]{Fig/20240903_124910.jpg}
    \vspace{0.4cm}
    \includegraphics[width=1\columnwidth]{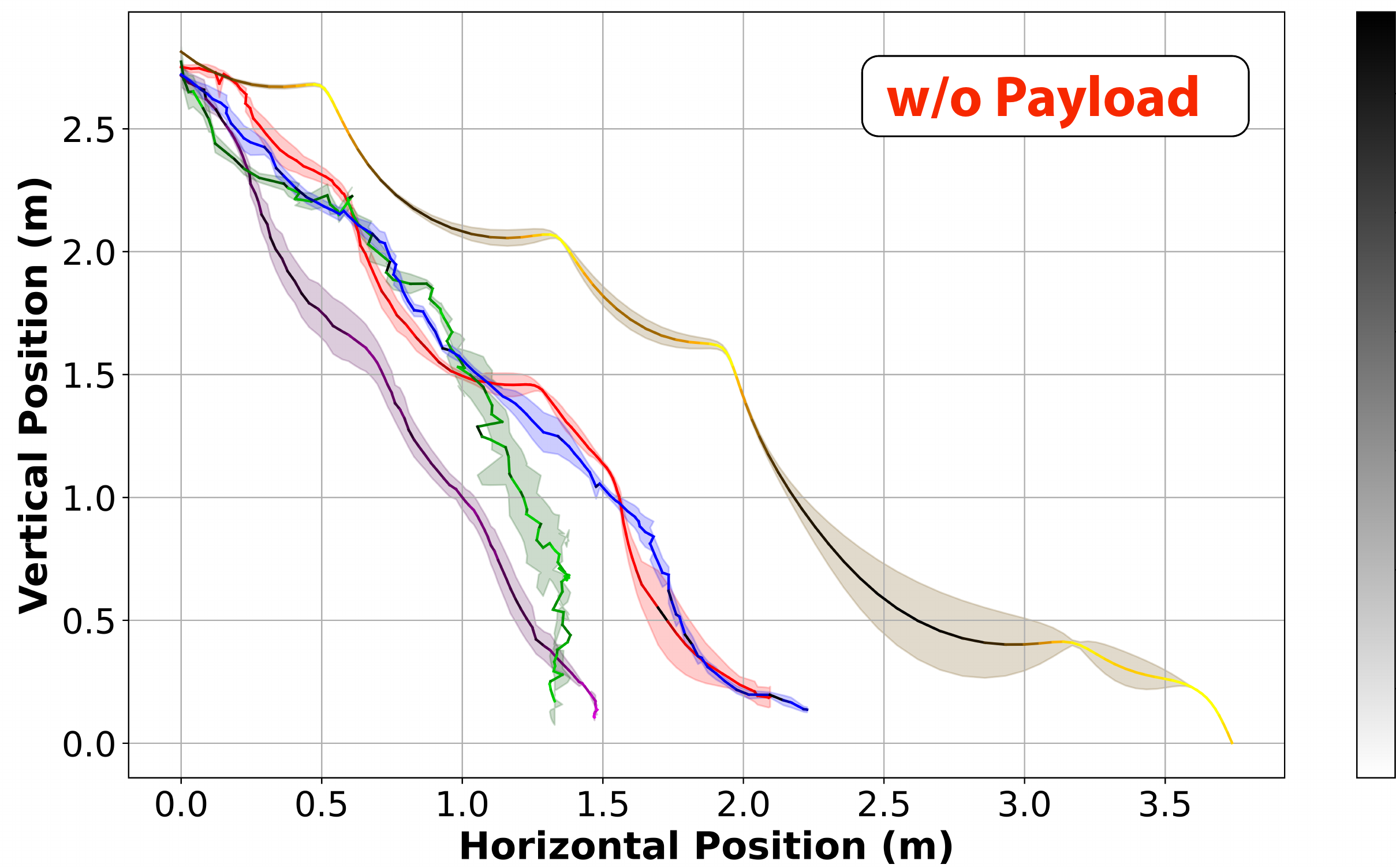}
    \includegraphics[width=1\columnwidth]{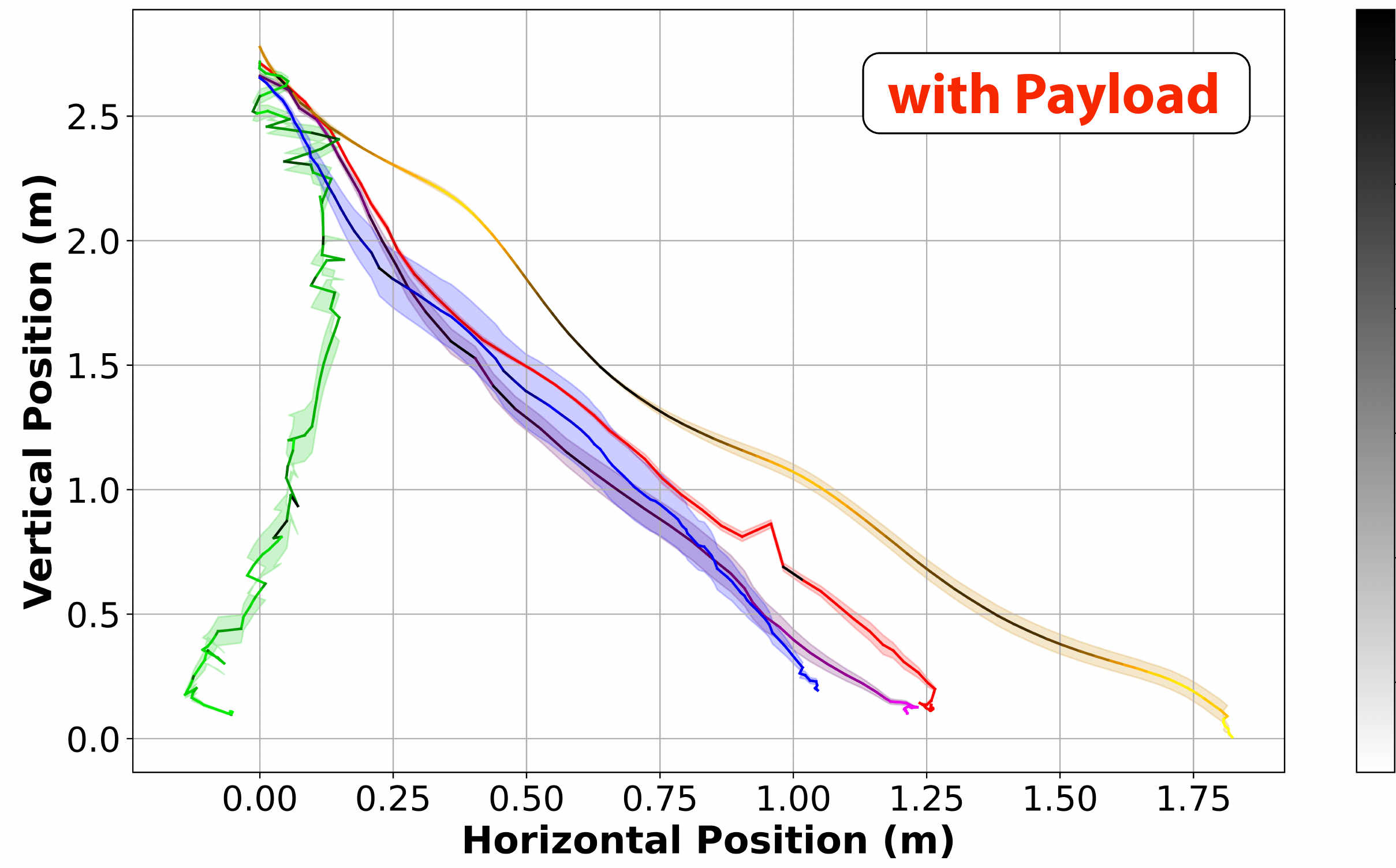}
    \caption{(Top row) Tumbler designs. (Middle) Tumbling tests of the above tumbler shapes without payload. (Bottom) Tumbling tests of the above tumbler shapes with a payload of 60 g. Right bar is an indication of the descend rate, ranging from 0 to 3 m/s.}
    \label{fig:combined_figures}
\end{figure}

We performed indoor tests, tracked by motion capture (Vicon, CA, USA), of designs 2-6, as outlined in \cref{fig:tumbler_designs}. Key parameters such as descent rate, position, speed, and rotations were analyzed. For consistency, all tumblers were released from the same height and pitch angle.
\cref{fig:combined_figures} illustrates the trajectories of the different shapes. Each tumbler was tested five times, and the plot represents the average trajectory from these tests, along with their standard deviation.

\label{sec:tests_results}
We noticed a distinct pattern in the trajectories resembling consecutive hyperbolic curves. The shift from one curve to the next corresponds to the moment when the tumbler flips over, losing the surface contact that initially generates lift. This flip leads to a sharp descent until the system regains enough speed to generate lift again, slowing down, and eventually flipping once more. This behavior is especially noticeable in the case of design 6, around the 1.5-meter mark in the vertical position, as seen in the first plot of \cref{fig:combined_figures}.
This pattern is clearly reflected in the speed plot, in which the descending rate of each system is illustrated using a color gradient, with darker shades indicating higher speeds and brighter shades representing lower speeds. The rate exhibits a sinusoidal profile corresponding to each flip in the tumbler's descent, peaking just before the tumbling flip, as seen in the case of design 6, depicted in orange in the figure. The speeds range from 0.8 m/s to 2 m/s, with a maximum peak recorded at approximately 2.5 m/s.
The gliding ratios of the tumblers range from 0.5 for design 5 to 1.5 for design 6. Gliding performance is strongly influenced by mass. In terms of shape, the dodecagon performs similarly to the circle when their masses are comparable, as demonstrated in designs 2 and 4.\\
A 60 g payload was attached to the tumbler to simulate the weight of the sensing unit, as illustrated in the second plot of \cref{fig:combined_figures}. The results are consistent with the previous tests. While the gliding ratios of the other shapes were reduced by a factor of 1.5 to 2, the square shape of design 3 was the only configuration that failed to initiate the tumbling motion entirely. In contrast, the lightweight dodecagon shape of design 6 continued to exhibit the highest gliding ratio and the most stable trajectory, with minimal deviation.
The initial pitch angle was another factor influencing the trajectory after deployment. Overall, the lower the pitching angle, i.e. vertical deployment, the faster the system reached the tumbling state.
Also in the payload cases, the rate shows a sinusoidal shape as seen in \cref{fig:combined_figures}, with an average increase of 20\% in descending rate with respect to the tumblers without payload.

\begin{figure}
\centering
\includegraphics[width=1\columnwidth]{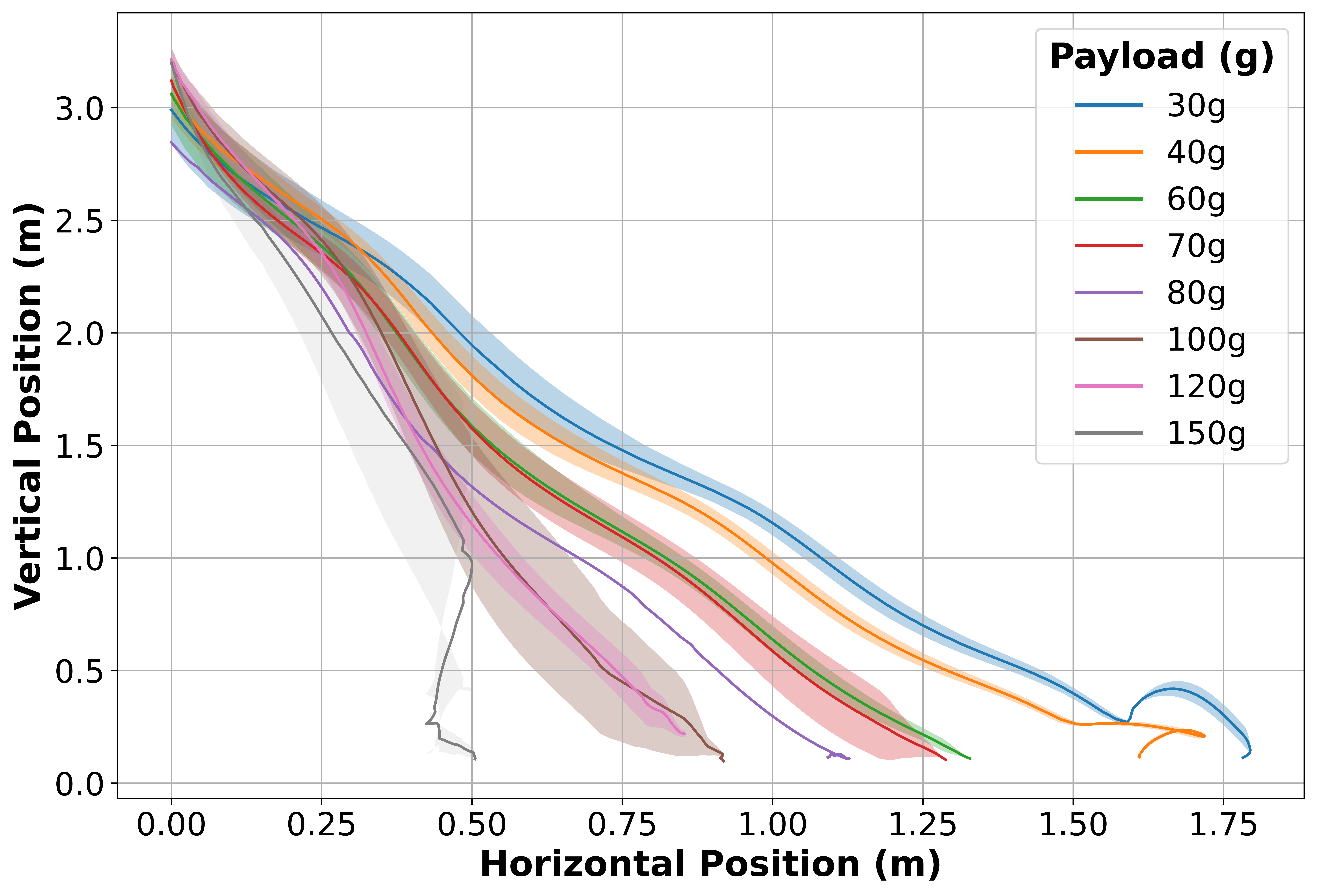}
\caption{Tumbling tests with increasing payload, starting with 30 g and up to 150 g.}
\label{fig:trajectory_payloads}
\end{figure}
To validate the tumbler designs under various task conditions using an alternative sensor unit, we conducted tests by equipping the tumbler with payloads ranging from 30 g to 150 g. For these experiments, design 5 was employed to ensure minimal bending even with higher payloads. As shown in \cref{fig:trajectory_payloads}, increasing the payload maintained the tumbling behavior and simply reduced the gliding ratio. This confirms the greater payload flexibility offered by tumbling systems over other deployment methods. Finally, with a payload higher than 120 g, the system could not tumble anymore, resulting in an uncontrollable and abrupt descent. This represents the maximum operational payload of the current system.

\subsection{Buoyancy tests}
\begin{figure}[b!]
\centering
\includegraphics[width=1\columnwidth]{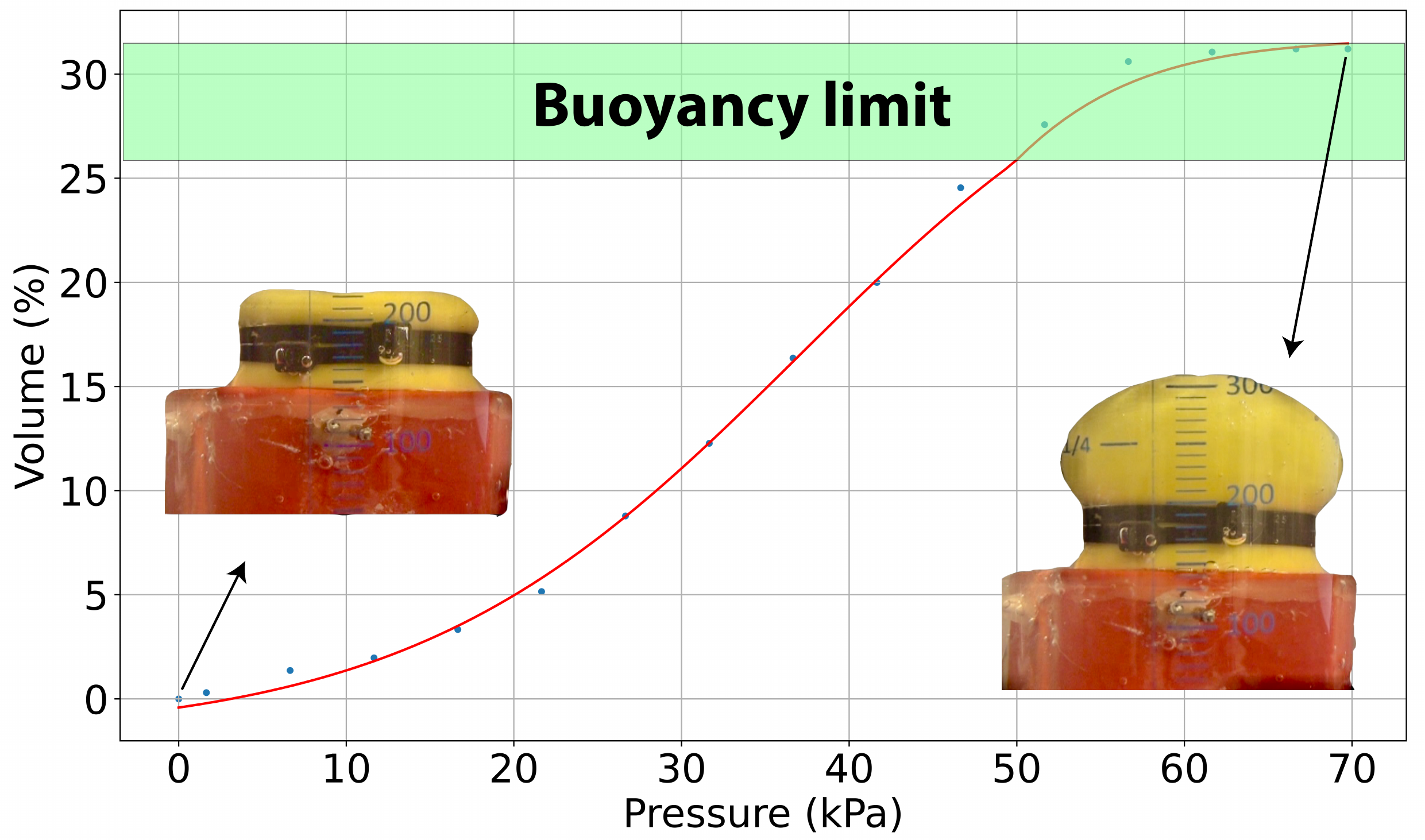}
\caption{Volume characterization was performed based on pressure differentials during buoyancy tests, with the red line fitting the volume-pressure relationship.}
\label{fig:buoyancy_tests}
\end{figure}

Underwater tests were conducted to validate the buoyancy system, as shown in \cref{fig:buoyancy_tests}. 
The volumetric percentage change was monitored with respect to the pressure generated by the chemical reaction. The system's behavior follows a hyperbolic curve, with a near-linear relationship between pressures of 20 and 50 kPa. The maximum pressure generated by the chemical reaction reached 164 kPa, resulting in a pressure differential of approximately 70 kPa with respect to the ambient environment. 
Under this pressure difference, the Benthic Sensing Pod is capable of achieving a volumetric change exceeding 30\%. The test indicates that a 5.7\% volume increase for positive buoyancy requires a pressure differential of approximately 20 kPa. Consequently, the maximum operational external pressure to achieve water resurfacing is around 50 kPa, as evident in the green area in  \cref{fig:buoyancy_tests}, corresponding to a depth of approximately 5 meters. While the pressure and temperature sensors can function at depths of up to 300 meters, the buoyancy system limits the maximum operational depth due to its role in recovery. This limitation arises not only from the quantity of chemicals used but also from challenges related to the integrity of the sealing structure, despite the silicon skin's ability to achieve greater volume change and withstand higher pressure.
In future tests, we aim to reach higher internal pressures in order to improve the maximal operational depth. This will mainly be achieved by increasing the amount of reactants and improving sealing between the casing and membrane. Further, a fully 3D printing design, including flexible materials for the membrane, will be explored.
\subsection{Outdoor tests}
Despite the extensive work carried out on the motion of thin discs falling in quiescent fluid \cite{Howison.2020, Field.1997}, very little is known about the motion of the developed tumblers falling through background turbulence \cite{Esteban2019}.
To this end, the tumblers have been tested in outdoor conditions and deployed on a small lake. Wind conditions were quite unfavorable with wind of 8 to 10 knots (4 to 5 m/s) and up to 15 knots of wind gusts. Nonetheless, the gripper was capable of ensuring a stable attachment to the tumbler, even in strong wind conditions, thanks to its simple yet robust design.
However, the large cross-section area of the tumbler introduced challenges in piloting the drone, reducing its roll capability and causing drift. The tumblers were tested both with and without payloads. Tumblers equipped with payloads demonstrated greater stability in windy conditions, ensuring system reliability during descent, with the glide ratio remaining comparable to indoor test results. 
\begin{figure}
\centering
\includegraphics[width=1\columnwidth]{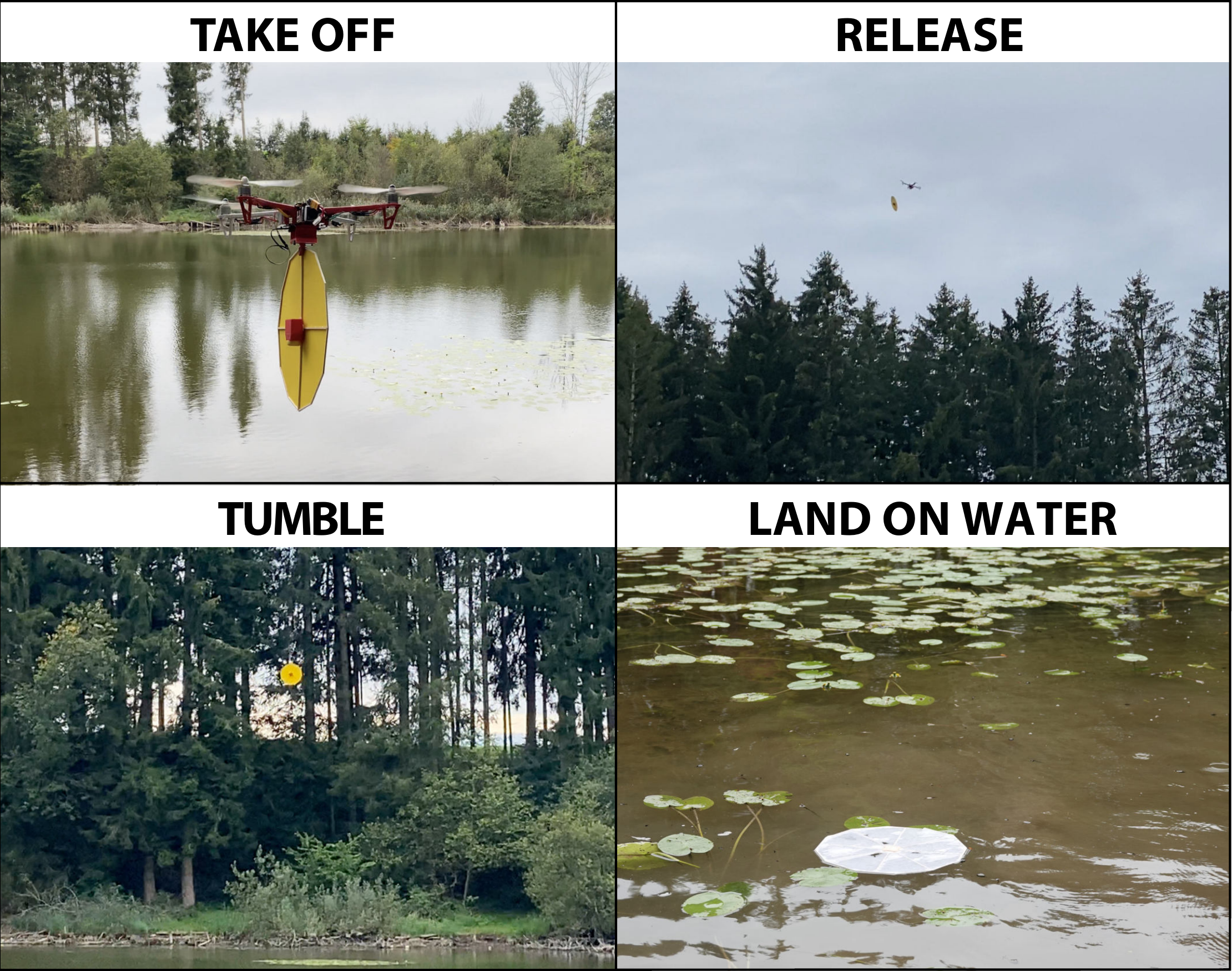}
\caption{Outdoor demonstration starting with the system take off (up-left), reaching an approximate height on 15 m and releasing the tumbler (up-right), reaching the tumbling state with the tumbler (down-left) and tumbler landing on water (down-right). Retrieval is accomplished by collecting the sensing unit from the shoreline.}
\label{fig:outdoor_mission}
\end{figure}
\section{Conclusions}
This work presents a novel framework for minimally-invasive benthic monitoring using a tumbler system deployed via drone. The tumbler descends toward the water at a maximum speed of 2 m/s, minimizing habitat disturbance. It carries a sensing pod equipped with temperature and pressure sensors. Upon water entry, the sensing unit detaches and descends to the benthic zone. After completing its mission, it ascends via a buoyancy mechanism that inflates a silicon membrane through a chemical reaction between citric acid and sodium bicarbonate.
A novel buoyancy mechanism was demonstrated, featuring a lightweight package and simple fabrication enabled by 3D printing. Future iterations will focus on increasing internal pressure by augmenting the quantity of reactants, thereby enhancing the system's maximum operational depth.
Various tumbler designs were tested, optimized for lightweight construction compatible with commercial quadrotors. Gliding ratios ranged from 0.5 to 1.5 without payload and reduced by half with a 60 g payload simulating the sensing unit. 
Increasing payloads were tested to evaluate the system's performance under additional weight, as its modular design supports the integration of sensors such as salinity, turbidity, pH, cameras, and microphones. The system supported payloads up to 120 g while maintaining stable tumbling motion. Outdoor tests using an F450 quadrotor demonstrated reliable deployment, with stability maintained under wind gusts up to 7.5 m/s, despite some influence on drone roll control. 
Tumbling systems offer significant potential for environmental sensing and sensor deployment, providing stability in windy conditions, low descent rates, and minimal drift.
This project highlights the potential for robotics in minimizing ecological disturbance during aquatic monitoring, providing a sustainable solution for environmental sensing missions.
Future work will involve deploying multiple tumblers in a swarm configuration, enabling spatial mapping through pod communication. Further environmental improvements could be achieved by replacing silicone components with biodegradable materials like gelatin and developing biodegradable sensors.



\bibliographystyle{IEEEtran}
\bibliography{references}

\end{document}